\documentclass{article}

\usepackage{iclr2027_conference,times}
\usepackage{amsmath}
\usepackage{amssymb}
\usepackage{booktabs}
\usepackage{multirow}
\usepackage{tabularx}
\usepackage{graphicx}
\usepackage{subcaption}
\usepackage{flafter}
\usepackage{placeins}
\usepackage{needspace}
\usepackage{microtype}
\usepackage{hyperref}
\usepackage{url}
\hypersetup{
  hidelinks,
  pdftitle={When History Fails to Become Experience: Action Calibration in Language Agents},
  pdfauthor={Jingyu Liu, Zhiwen Wang, Yuxin Jing, Huanyu Zhou, Yong Liu}
}

\AddToHook{cmd/section/before}{\Needspace{5\baselineskip}}
\AddToHook{cmd/subsection/before}{\Needspace{4\baselineskip}}

\title{When History Fails to Become Experience:\\
Action Calibration in Language Agents}
\author{\parbox{0.96\textwidth}{\centering
\textbf{Jingyu Liu}\textsuperscript{1}\quad
\textbf{Zhiwen Wang}\textsuperscript{2}\quad
\textbf{Yuxin Jing}\textsuperscript{2}\quad
\textbf{Huanyu Zhou}\textsuperscript{2}\quad
\textbf{Yong Liu}\textsuperscript{1,3,4}$^{\dagger}$\\[3pt]
{\normalfont\small
$^1$ Gaoling School of Artificial Intelligence, Renmin University of China, Beijing, China\\
$^2$ ByteDance\\
$^3$ Beijing Key Laboratory of Research on Large Models and Intelligent Governance\\
$^4$ Engineering Research Center of Next-Generation Intelligent Search and Recommendation, MOE\\
\texttt{liujy1016@ruc.edu.cn}\quad
$^{\dagger}$\texttt{liuyonggsai@ruc.edu.cn}}
}}
\date{}
\iclrpreprintcopy

\begin{document}

\maketitle

\begin{abstract}

Language agents should draw on prior attempts and environmental feedback to improve subsequent decisions within the same task.
However, providing additional interaction history can sometimes reduce task success, suggesting that agents do not
consistently use this information effectively. To investigate this limitation, we examine how
agents use history. We find that history improves task completion overall, yet much
of this benefit persists even when past actions are shuffled. Disrupting the correspondence
between actions and observations causes only a modest decline in task success.
We therefore hypothesize that agents do not reliably connect past actions with their outcomes when
deciding how to proceed. To test this hypothesis, we explicitly label each returned observation as
the outcome of the preceding action. This simple annotation improves task success and reduces next-action
repetition without introducing new environmental information. Building on this insight, we introduce a learned calibrator
that explicitly reassesses past actions and selectively records experience to guide subsequent decisions,
improving task success beyond outcome labeling alone.

\end{abstract}

\section{Introduction}
\label{sec:introduction}

A line of work on language agents and self-evolving systems distills past trajectories into
reusable lessons and guidelines for subsequent tasks \citep{ding2026apex,xu2026hyperskill}. Within an ongoing task,
however, the agent's own actions and environmental feedback are already available as an
\emph{interaction history} \citep{yao2023react}. These attempts can reveal what worked, what
failed, and what needs to change. This history therefore provides task-specific evidence that
should help the model make better decisions.

However, interaction history can also reinforce unsuccessful behavior. Prior work links conversational
inertia to imitation of agents’ own responses \citep{wan2026inertia},
and finds that failed action strings can sometimes outweigh corrective feedback
in small models \citep{gumaan2026feedback}. The same history that encourages an agent to
repeat an unsuccessful action also contains feedback indicating its failure.
This raises the question of how agents use that evidence within an ongoing task.

\begin{figure}[t]
  \centering
  \includegraphics[width=\linewidth,trim=0 10bp 0 10bp,clip]{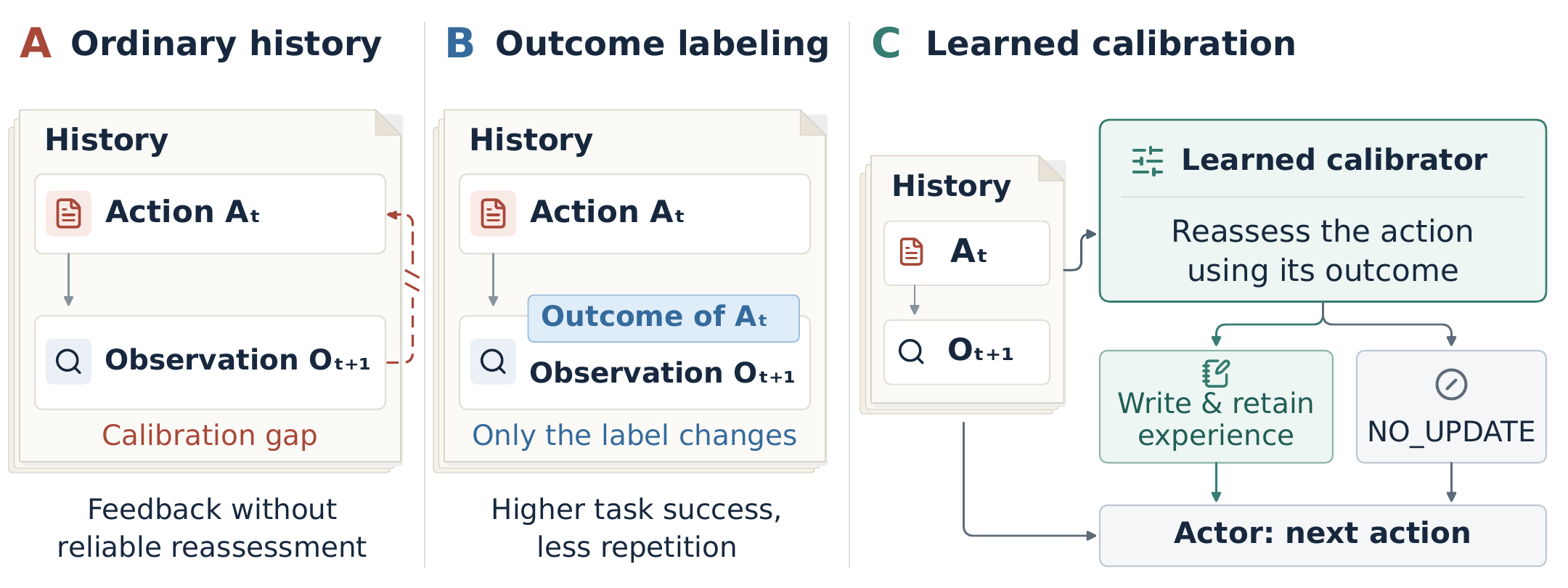}
  \caption{Action calibration within an ongoing task.
  (A) Ordinary history simply retains the observation returned by the environment.
  (B) Outcome annotates the observation as the outcome of the previous action.
  (C) A calibrator selectively retains experience to guide later decisions.}
  \label{fig:action_calibration_overview}
\end{figure}

We therefore test whether additional history helps the agent.
We restore an early environment state from a failure trajectory
and compare continuations with the original prefix alone or with
additional future actions and feedback from the same trajectory, provided as reference history.
The reference history is expected to help the model better understand the environment and identify what
might help and what might hurt.
But overall recovery is lower with the additional reference
history: in this setting, more attempts and feedback make it harder for the agent to complete the task
instead of helping it understand the environment.

To better understand how models use interaction history, we shuffle past actions while
leaving observations in their original order, disrupting the original action--outcome
pairings. Overall task success nevertheless remains close to that with the original history.
These results suggest that models make limited use of the original action--outcome pairings
in ordinary history. Together with the recovery results, this motivates our hypothesis that ordinary history does not reliably elicit
\emph{action calibration}: using an action's outcome to decide whether an earlier approach
should be continued, adjusted, or abandoned.

One possibility is that the model treats an observation primarily as context for the action taken in
response, rather than as the outcome of the previous action. So it fails to calibrate the action based on its outcome.
To test this hypothesis,
we annotate each returned observation as the
outcome of the action that produced it (outcome of action A: new observation). This change draws attention to what a past
action actually achieved, without supplying new environmental information or an additional
judgment about whether the action was appropriate. This annotation improves overall task success
and reduces action repetition. These gains further support the hypothesis that agents underuse
existing action--outcome feedback in ordinary history.

Outcome labels leave the agent to implicitly decide what to learn from each attempt. We therefore ask
the model to explicitly decide whether calibration is needed and perform calibration by
summarizing and recording \emph{experience}
from its interaction history. Yet the expected improvement does not follow: retaining
self-generated experiences throughout a task lowers overall success relative to ordinary history.
Inspecting the trajectories reveals that some experiences merely restate available feedback,
while others even introduce unsupported judgments.

These results motivate learning which experiences to retain. Their effects on next-action
quality across multiple actors provide supervision for a calibrator that learns when an update
is warranted and what experience to write. The learned calibrator improves task completion on
held-out tasks over both ordinary history and outcome labeling. Figure~\ref{fig:action_calibration_overview}
summarizes the calibration gap, outcome labeling, and learned calibration.

Our contributions are fourfold:
\begin{enumerate}
  \item We show that additional reference history can reduce recovery.
  \item History interventions further
  reveal that historical information helps task completion, and much of this benefit persists
  when historical action strings are shuffled.
  \item We show that annotating observations as action outcomes improves task success and
  reduces next-action repetition on fixed histories.
  \item We train a calibrator to select and generate experiences, improving task completion on held-out tasks.
\end{enumerate}

\FloatBarrier
\section{Related Work}
\label{sec:related_work}

\paragraph{Summarizing and using experience.}
Agents extract reusable guidance from past interactions, ranging from verbal reflections across
trials \citep{shinn2023reflexion} to procedural skills from trajectory memories
\citep{ding2026apex}. Learning also shapes how this guidance is managed and reused for future tasks: agents can
evolve memory operations and learn which to apply \citep{zhang2026memskill}, or use environmental
feedback to update experience utility and guide retrieval \citep{zhang2026memrl}. These methods explicitly
construct guidance from previous trajectories.
We study whether agents use feedback already in an ongoing task's history to revise earlier
approaches.

\paragraph{Negative effects of interaction history.}
Conversational and contextual inertia can reinforce prior responses and reasoning despite new
constraints \citep{wan2026inertia,chen2026contextualinertia}.
Failed tool calls can encourage repetition despite error feedback
\citep{gumaan2026feedback}. EVU \citep{wang2026beliefinertia} identifies \emph{belief inertia}: observations
that contradict an agent's expectations may fail to update its internal
understanding of the environment in small models. However, these studies do not examine how
the negative effects of history interact with the additional information it provides, or the reason why current models fail to
take advantage of feedback.

\section{More History Does Not Ensure Better Recovery}
\label{sec:motivation}
\label{sec:failure_conditioned_raw_recovery}

As interaction proceeds, an agent accumulates more information about the environment.
This information should help the agent
avoid ineffective choices and reassess its approach. We test this expectation by restoring an
early state of each unsuccessful trajectory and then providing the model with additional future interaction
history as a reference to show how the extra information affects performance.

Specifically, for a source trajectory of length $L$, we restore the environment to
$\lfloor L/3\rfloor$, and instruct the model to continue the task with the following two settings.
\begin{enumerate}
  \item \textbf{Raw} provides only the action--observation history through that point.
  \item \textbf{Reference} additionally provides the recorded segment from
$\lfloor L/3\rfloor$ to $\lfloor2L/3\rfloor$ in a separate reference block, explicitly marked
as not executed in the restored environment.

\end{enumerate}

For example, if a source trajectory fails after
30 actions, both continuations start from the state after action 10; the second also receives
actions 11--20 and their outcomes as reference information. The actor receives the actual
current observation and available actions, together with any newly executed continuation
history. Both conditions use the source actor and allow up to 30 new actions. Recovery means
terminal task success within this budget.

\begin{figure}[!htbp]
  \centering
  \includegraphics[width=\linewidth,trim=0 2bp 0 2bp,clip]{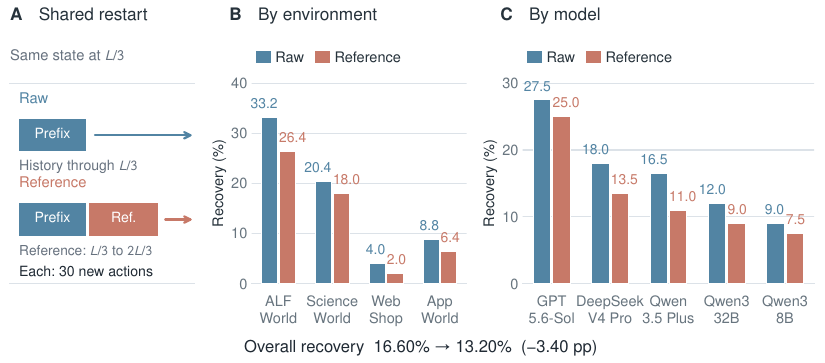}
  \caption{Recovery from the same restored state. For a failed trajectory of length $L$, Panel A
  contrasts Raw ($L/3$ history) with Reference (adding the recorded $L/3$--$2L/3$ segment,
  not executed in the restored environment). Both conditions resume at $L/3$ with a budget of
  30 new actions, using the same 1,000 source trajectories. Panels B--C pool recovery by
  environment and model, respectively.}
  \label{fig:failure_conditioned_raw_recovery}
\end{figure}

We evaluate GPT-5.6-Sol, DeepSeek-V4-Pro, Qwen3.5-Plus, Qwen3-32B, and Qwen3-8B in ALFWorld,
ScienceWorld, WebShop, and AppWorld
\citep{shridhar2021alfworld,wang2022scienceworld,yao2022webshop,trivedi2024appworld}.

We can observe from Figure~\ref{fig:failure_conditioned_raw_recovery} that adding the reference history lowers recovery,
despite providing more attempts and feedback from the same task. Recovery is lower in all
four environment aggregates and all five model aggregates.
Appendix~\ref{app:reference_recovery} gives the prompt format.
The same pattern holds with separate user and assistant messages: adding reference history
reduces recovery from 15.8\% to 11.4\% (Appendix~\ref{app:native_chat_continuation}).

\begin{figure}[!htbp]
  \centering
  \includegraphics[width=\linewidth,trim=0 1bp 0 1bp,clip]{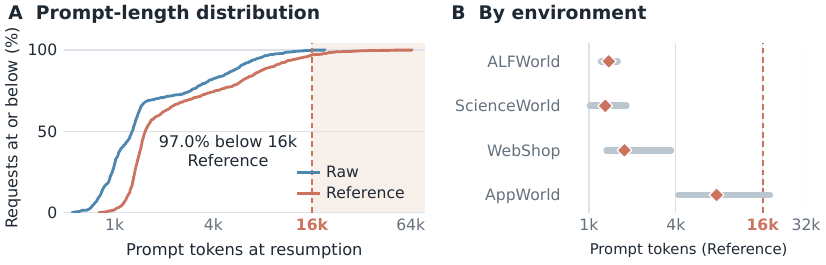}
  \caption{Prompt lengths at the first request after resumption, for 1,000 source trajectories.
  Panel A compares Raw with Reference; Panel B shows Reference medians (diamonds) and
  10th--90th percentiles (lines) by environment. Both horizontal axes are logarithmic.
  The dashed line at 16,384 tokens is a visual reference, not a model context limit.}
  \label{fig:nested_checkpoint_prompt_lengths}
  \label{fig:nested_checkpoint_prompt_length_cdf}
  \label{fig:nested_checkpoint_prompt_length_environment}
\end{figure}

Because adding history also lengthens the input, we examine prompt lengths with and without
reference history. As shown in Figure~\ref{fig:nested_checkpoint_prompt_lengths}, prompts with
additional reference history have a median length of approximately 1.6k tokens. We retain only
historical observations and actions, omit intermediate reasoning, and format the history as in
\citet{GiGPO}. This length is way below the context capacity of current models, especially
state-of-the-art models such as GPT-5.6-Sol, showing that the performance decline cannot be
explained solely by context length.

\FloatBarrier
\section{How Agents Use Their History}
\label{sec:action_history_utilization}
\label{sec:recent3_information}

Additional reference history does not, by itself, yield better recovery from these
unsuccessful trajectories. We therefore ask how much agents benefit from retaining historical
information and how much their performance depends on preserving its original action--outcome
organization.

We compare ordinary history (\textbf{Raw}) with four interventions in full rollouts from the
initial task state (Table~\ref{tab:history_rendering_example}).
\textbf{Recent-3-Only} removes all but the three latest
history entries; \textbf{Observation-Only} removes historical action strings but retains all
observations in order. \textbf{Shuffle-All} retains both kinds of content but randomly reassigns
actions among all historical action positions. \textbf{Shuffle-Old} instead preserves the three latest entries
and reassigns older actions only among earlier positions. Both shuffle conditions keep observations
in chronological order while changing the placement of action strings and their original
correspondences with observations.

\begin{table}[!htbp]
  \centering
  \caption{Five history views at the same point in an interaction. $O_t$ is an observation;
  action $A_t$ responds to it and produces $O_{t+1}$. Orange denotes reassigned actions;
  the permutations are illustrative.}
  \label{tab:history_rendering_example}
  \setlength{\tabcolsep}{4pt}
  \begin{tabular}{@{}lll@{}}
    \toprule
    \textbf{Condition} & \textbf{Earlier history} & \textbf{Recent history} \\
    \midrule
    Raw & $(O_1,A_1),(O_2,A_2),(O_3,A_3)$ & $(O_4,A_4),(O_5,A_5),(O_6,A_6)$ \\
    Recent-3-Only & \textit{Entire entries omitted} & $(O_4,A_4),(O_5,A_5),(O_6,A_6)$ \\
    Observation-Only & $O_1,O_2,O_3$ & $O_4,O_5,O_6$ \\
    Shuffle-All & $(O_1,\textcolor{orange!85!black}{A_4}),(O_2,\textcolor{orange!85!black}{A_5}),(O_3,\textcolor{orange!85!black}{A_6})$ & $(O_4,\textcolor{orange!85!black}{A_1}),(O_5,\textcolor{orange!85!black}{A_2}),(O_6,\textcolor{orange!85!black}{A_3})$ \\
    Shuffle-Old & $(O_1,\textcolor{orange!85!black}{A_2}),(O_2,\textcolor{orange!85!black}{A_3}),(O_3,\textcolor{orange!85!black}{A_1})$ & $(O_4,A_4),(O_5,A_5),(O_6,A_6)$ \\
    \midrule
    \multicolumn{3}{c}{Current observation $O_7$ is retained in all five conditions.} \\
    \bottomrule
  \end{tabular}
\end{table}

Each of the five history conditions contains 500 tasks per model--environment cell,
or 10,000 episodes per condition across the five actors and four
environments.

Removing earlier history substantially lowers overall success
(Table~\ref{tab:history_intervention_marginals}). Recent context is therefore not generally
a substitute for accumulated information. Observation-Only retains much of Raw's success
and outperforms Recent-3-Only overall: historical observations preserve useful task information
even when the actions that accompanied them are omitted. 
We show full results in Appendix~\ref{app:action_history_full}.

\begin{table}[!htbp]
  \centering
  \caption{Task success (\%) under history interventions. The two blocks summarize the same
  evaluation grid: 2,500 tasks per
  environment, 2,000 per model, and 10,000 overall, per condition.
  Table~\ref{tab:action_history_cells} gives the full breakdown.}
  \label{tab:history_intervention_marginals}
  \begin{tabular}{@{}lccccc@{}}
\toprule
Group & \multicolumn{1}{c}{Raw} &
\multicolumn{1}{c}{\begin{tabular}[c]{@{}c@{}}Recent-3-\\Only\end{tabular}} &
\multicolumn{1}{c}{\begin{tabular}[c]{@{}c@{}}Observation-\\Only\end{tabular}} &
\multicolumn{1}{c}{Shuffle-All} & \multicolumn{1}{c@{}}{Shuffle-Old} \\
\midrule
\multicolumn{6}{@{}l}{\textit{By environment}} \\
\addlinespace[0.1em]
ALFWorld & 67.08 & 52.44 & 61.88 & 64.64 & 68.08 \\
ScienceWorld & 39.12 & 31.88 & 35.88 & 34.84 & 38.56 \\
WebShop & 27.72 & 27.28 & 20.68 & 25.44 & 27.88 \\
AppWorld & 51.64 & 32.36 & 45.28 & 50.44 & 52.48 \\
\addlinespace[0.3em]
\multicolumn{6}{@{}l}{\textit{By model}} \\
\addlinespace[0.1em]
GPT-5.6-Sol & 73.30 & 60.20 & 68.85 & 70.75 & 73.45 \\
DeepSeek-V4-Pro & 63.05 & 46.40 & 56.00 & 57.65 & 62.35 \\
Qwen3.5-Plus & 48.30 & 37.15 & 41.10 & 48.00 & 49.80 \\
Qwen3-32B & 29.05 & 22.95 & 22.30 & 27.20 & 29.80 \\
Qwen3-8B & 18.25 & 13.25 & 16.40 & 15.60 & 18.35 \\
\midrule
\textbf{Overall} & \textbf{46.39} & \textbf{35.99} & \textbf{40.93} & \textbf{43.84} & \textbf{46.75} \\
\bottomrule
\end{tabular}

\end{table}
\FloatBarrier

Shuffle-All retains most of Raw's performance, with an overall decrease of only 2.55 percentage
points (Table~\ref{tab:history_intervention_marginals}). Task success is thus only modestly
sensitive to disrupting the original action--outcome pairings. Shuffle-Old improves over
Shuffle-All and stays close to Raw: when recent interactions remain intact, preserving earlier
pairings provides little additional overall benefit. Preserving recent interactions may also
make the immediate context more coherent and familiar, so this gain need not reflect action
calibration.

Historical content remains useful despite disruption of its original action--outcome
organization. Together with poorer recovery from additional reference history
(Section~\ref{sec:motivation}), this motivates our hypothesis that ordinary history does not
reliably elicit \emph{action calibration}. Agents may extract useful information without
sufficiently using outcomes to reassess whether earlier approaches should be continued,
adjusted, or abandoned.

\Needspace{8\baselineskip}
\section{Turning Recorded Feedback into Useful Experience}
\label{sec:explicit_calibration}

\subsection{Making Action Outcomes Explicit}
\label{sec:fixed_history_repetition}

To test this hypothesis, we ask whether explicitly identifying an action's outcome helps the
agent use feedback already in its history to reassess previous attempts.

\begin{figure}[!htbp]
  \centering
  \includegraphics[width=\linewidth,trim=0 0 0 2bp,clip]{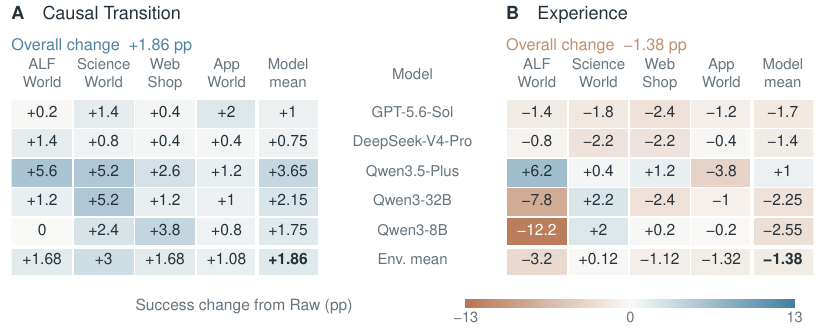}
  \caption{Task-success changes from the matching model--environment Raw baseline, in percentage
  points (500 tasks per cell). Both panels use the same color scale. The final row and column
  give environment and model means; the lower-right cell gives the pooled change over
  10,000 episodes per condition.}
  \label{fig:explicit_interfaces_success}
  \label{fig:explicit_interface_success_cells}
  \label{fig:causal_diagnostics}
  \label{fig:explicit_outcomes_n500_success}
\end{figure}

We therefore use \textbf{Causal Transition} to make each action's outcome explicit. We label
each returned observation as the outcome of its preceding action:
$[\text{Outcome of }A_{t-1}:O_t,A_t]$, followed by
$[\text{Outcome of }A_t:O_{t+1},A_{t+1}]$. The underlying observations and their order are unchanged.
Figure~\ref{fig:explicit_interfaces_success} shows that Causal Transition improves overall task success over Raw.
Gains appear in most model--environment combinations and in every model-level and
environment-level aggregate.
Explicitly reminding the actor to recheck its history yields a similar overall success
rate to Causal Transition (Appendix~\ref{app:history_check}).

We hypothesize that this difficulty arises in part from the next-action objective.
Language agents are typically asked to answer a question or choose the next action from the
current observation. This objective may encourage them to respond directly to the latest
observation without relating it back to the action that produced it.

To test whether outcome annotation changes action reuse, we measure next-action repetition
on fixed histories.
We use exact-action repetition as a directly measurable indicator of action reuse
\citep{gumaan2026feedback}. We select histories ending in two identical actions, filtered for
feedback that does not support another repetition (judged by GPT-5.6-Sol). The source actor generates one next action
from each prefix under Raw and Causal, and we check whether the actor repeats the action
(Appendix~\ref{app:loop2_same_source_repetition}).

\begin{figure}[!htbp]
  \centering
  \includegraphics[width=\linewidth]{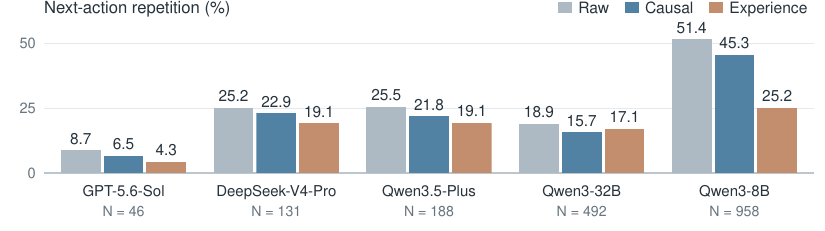}
  \caption{Probability of repeating the same action a third time. Each model receives the same
  $N$ histories in all three conditions, ending in two identical actions and screened for
  feedback that does not support another repetition. Lower is better.}
  \label{fig:explicit_interfaces_repetition}
  \label{fig:calibration_repetition_main}
\end{figure}

Figure~\ref{fig:explicit_interfaces_repetition} shows lower next-action repetition under Causal
for all five models. The annotation supplies neither new evidence nor advice about which
action to take, yet it changes action reuse on the same preceding history. Together with the
rollout gains, this result shows that identifying existing feedback as an action's outcome
can affect both the next decision and task completion. These results support our hypothesis
that ordinary history does not fully elicit outcome-based reassessment of past actions.

\subsection{From Outcome Labels to Explicit Experience}
\label{sec:self_persistent_calibration}

Annotating an observation as an action's outcome tells the actor what happened, but leaves it to
decide what to learn from that result. Since outcome labeling already helps, we ask whether
explicitly stating that lesson as an experience provides further benefit.

Under \textbf{Experience}, we retain the Raw history format and append actor-generated
experiences to the history. After each completed nonterminal interaction, we make a separate review call to
the same actor, denoted $C_{\mathrm{actor}}$. Its input $x_t$ contains the task and raw history
through action $A_t$ and its returned observation $O_{t+1}$. We instruct the model to generate
an experience based on the given action and returned observation when needed.
The experience-generation prompt is provided in Appendix~\ref{app:experience_review_prompt}.

\Needspace{4\baselineskip}
The call produces either a
task-relevant experience $E_t$ or \texttt{NO\_UPDATE}:
\begin{equation}
  y_t=C_{\mathrm{actor}}(x_t)\in
  \bigl\{\texttt{NO\_UPDATE},\;(\texttt{CALIBRATE},E_t)\bigr\}.
  \label{eq:calibrator_decision}
\end{equation}

A generated experience $E_t$ remains in the actor's history for subsequent decisions;
\texttt{NO\_UPDATE} leaves that history unchanged. This review asks the actor to use
environmental feedback to explicitly decide whether an earlier approach should be continued,
adjusted, or abandoned.
Appendix~\ref{app:experience_decision_rates} reports how often each actor generates an
experience or returns \texttt{NO\_UPDATE} in each environment.

However, experience does not deliver the expected gain in task completion. In full rollouts, its overall
success rate is lower than both Raw and Causal (Figure~\ref{fig:explicit_interfaces_success}).
It reduces next-action repetition relative to Raw across all five models in the
fixed-history probe, with a lower pooled rate than Causal
(Figure~\ref{fig:explicit_interfaces_repetition}). But this reduction in repetition does not
translate into improved task success.
These results are consistent with prior findings that generated reflections can be ineffective
or counterproductive for interactive agents~\citep{li2025dora,lippmann2025positive}.

\begin{figure}[!htbp]
  \centering
  \includegraphics[width=\linewidth,trim=0 2bp 0 2bp,clip]{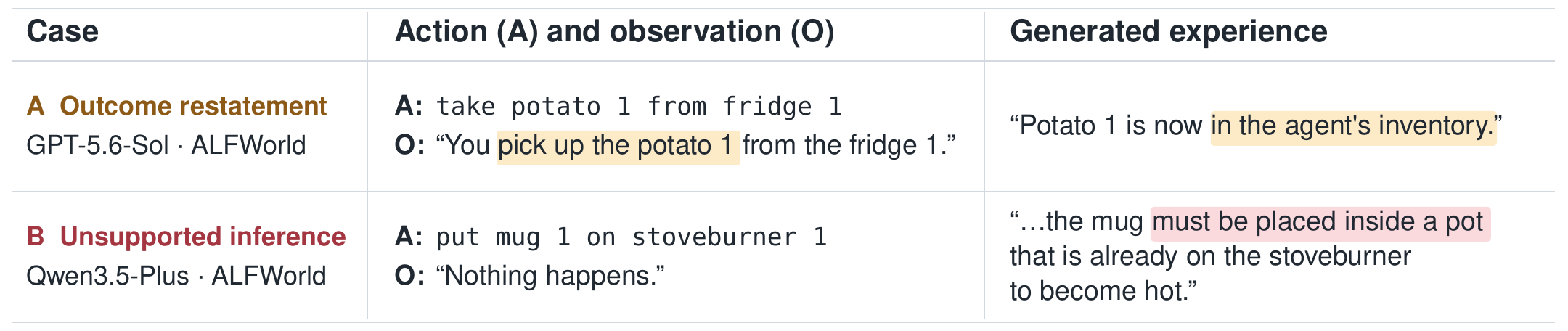}
  \caption{An outcome restatement and an unsupported inference in actor-generated experience.
  Highlighting marks the repeated fact or unsupported rule. Quotations come from real
  ALFWorld trajectories; the second experience is excerpted.}
  \label{fig:experience_quality_cases}
\end{figure}

To understand what causes the performance degradation, we inspect the experiences generated during
full rollouts. We find that current models can produce poor experiences: some simply repeat the
observation, while others introduce unsupported claims. Figure~\ref{fig:experience_quality_cases}
shows examples.
The GPT-5.6-Sol experience simply restates an inventory change already reported in
the observation. The Qwen3.5-Plus experience introduces a claim unsupported by the environment
feedback.
Appendix~\ref{app:self_persistent_cases} reports further examples.
Appendix~\ref{app:experience_next_action_quality} further shows that generated experiences can
degrade next-action quality, including experiences generated by GPT-5.6-Sol.

\Needspace{5\baselineskip}
Causal labels existing feedback as an action's outcome. Experience goes further by adding
the actor's interpretation and retaining it as guidance for later decisions. Because this
guidance persists, redundant or unsupported experiences can affect more than the immediately
following action. The next question is therefore which experiences should be retained.

\section{Learning When and How to Calibrate Actions}
\label{sec:outcome_triggered_calibration}

\subsection{Constructing Labels from Paired Next-Action Comparisons}

The Experience baseline already lets the actor skip an update, yet it can retain redundant or
unsupported guidance. Learning which experiences to retain therefore calls for supervision
grounded in their effects on decisions. We use changes in next-action quality across multiple
actors as a signal for when to retain an experience and what to write.

We sample 10,000 real decision contexts evenly across the five source actors and four
environments, using rollouts on tasks disjoint from the final evaluation. GPT-5.6-Sol reviews
each context and returns either \texttt{NO\_UPDATE} or a candidate experience $E$. A direct
\texttt{NO\_UPDATE} response becomes a \texttt{NO\_UPDATE} training target. Then,
a GPT judge screens candidate experiences to exclude unsupported claims and
experiences that merely repeat the observation. If a candidate fails this screening, its context
is retained with a \texttt{NO\_UPDATE} training target.

Only candidates that pass this content screening are evaluated with all five actors.
Each actor generates a next action twice from the same decision context: once
from the original prompt and once with $E$ added. The two generations form a within-actor pair
that differs only in whether the candidate experience is present. In a separate judging call,
GPT-5.6-Sol compares each pair and classifies the change in next-action quality as improved,
unchanged, or degraded. The judge considers whether the actions respect execution constraints,
plausibly advance the task, and avoid attempts already shown to be unproductive by the recorded
feedback.

Let $n_{\uparrow}(E)$ and $n_{\downarrow}(E)$ be the numbers of actors whose next-action quality
improves or degrades with $E$. We define the cross-actor quality margin as
\begin{equation}
  q(E)=n_{\uparrow}(E)-n_{\downarrow}(E).
  \label{eq:cross_actor_quality_margin}
\end{equation}
If $q(E)>0$ or $n_{\downarrow}(E)=0$, the training target is \texttt{CALIBRATE} followed by $E$.
Otherwise, the target is \texttt{NO\_UPDATE}.
Thus, a candidate experience is retained when
improvements outnumber degradations or when no actor's next-action quality degrades.
Candidates that leave next-action quality unchanged for all actors are retained because
their guidance may become useful later in the trajectory.

We fine-tune a Qwen3-8B calibrator on these prompt--target pairs using supervised fine-tuning
for 3 epochs with learning rate 1e-5.
Each target contains either \texttt{NO\_UPDATE} or \texttt{CALIBRATE} followed by the
selected experience. The model therefore learns both whether to write an update and what to write.
During rollout, the frozen calibrator reviews each completed nonterminal interaction and either
adds an experience to the history or returns \texttt{NO\_UPDATE} without changing it. Generated
experiences persist in the history, while the original actor continues to generate actions.
Appendix~\ref{app:prompt_templates} records the core interfaces.

\subsection{Task Completion Evaluation}
\label{sec:full_rollout_evaluation}

We compare the calibrator with Raw and Causal Transition using the same five actors
and four environments, with 500 tasks per model--environment cell. The rollout tasks are
distinct from the calibrator's training tasks.

\begin{figure}[!htbp]
  \centering
  \includegraphics[width=\linewidth,trim=0 5bp 0 1bp,clip]{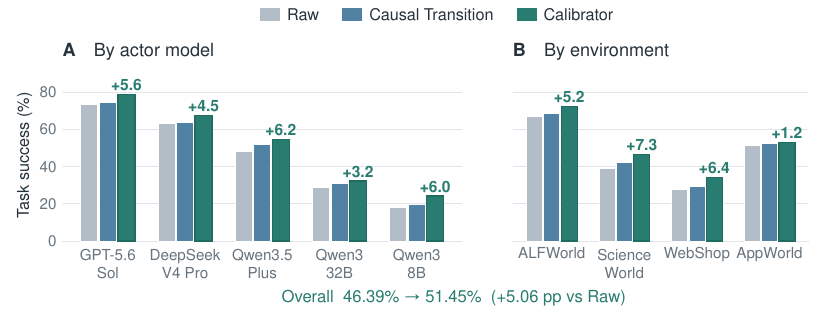}
  \caption{Full-rollout success (\%) for Raw, Causal Transition, and Calibrator.
  Panel A pools 2,000 tasks per actor; Panel B pools 2,500 per environment
  (500 per model--environment cell). Labels show Calibrator gains over Raw in percentage points;
  Overall pools all 10,000 tasks per condition.}
  \label{fig:outcome_triggered_calibration_success}
\end{figure}

Figure~\ref{fig:outcome_triggered_calibration_success} shows that the calibrator improves
task success over Raw in every model-level and environment-level aggregate.
Qwen3.5-Plus shows the largest performance gain and also benefits most from labeling
observations as action outcomes (Figure~\ref{fig:explicit_interfaces_success}). We hypothesize
that these larger gains reflect weaker action-calibration ability than in the other models.

A calibrator trained with content screening alone, without paired next-action comparisons
across actors, achieves lower overall success than the full pipeline
(Appendix~\ref{app:calibrator_content_ablation}). This supports assessing experiences through
their effects on actor decisions.

\subsection{Failure Recovery}

To test whether our method and outcome annotation help the model use historical information,
we return to the same-state recovery comparison in Section~\ref{sec:motivation}.

Can changing how feedback is presented and reviewed improve recovery when reference history
is available? Alongside Raw and Reference, we evaluate three reference-augmented conditions.
\textbf{Causal + Reference} labels observations as action outcomes in both the actual history
and the reference segment. \textbf{Experience} generates an experience once at resumption and
retains it, while \textbf{Calibrator} uses the learned update rule during continuation. All five
conditions share the source trajectories, restored $L/3$ state, actor, and 30-new-action budget.

All three interventions raise pooled recovery above both unmodified Reference and the Raw
prefix-only baseline, with Calibrator achieving the highest recovery
(Figure~\ref{fig:continuation_reference_conditions}). 
The recovery gain from Experience shows that reviewing failed histories can produce useful
guidance for subsequent decisions. In general full rollouts, however, our trajectory inspection
also identifies redundant or unsupported experiences that can misguide later decisions and lead to
worse overall performance.

Thus, the recovery loss from supplying
unannotated reference history can be reversed by labeling action outcomes or adding explicit
guidance. This shows that these methods can effectively suppress the negative effects of
history and take advantage of the extra information.
Appendix~\ref{app:reference_recovery} provides further details.

\Needspace{8\baselineskip}
\subsection{Integrating Calibration into the Actor}
\label{sec:joint_calibration_actor}

Can the actor learn to generate this guidance alongside its action, without a separate
calibration call? We further integrate the calibrator into the actor by training two
Qwen3-8B variants. The action-only baseline is
trained on successful trajectories generated by GPT-5.6-Sol.
The joint variant uses successful trajectories generated by GPT-5.6-Sol with our calibrator
and combines calibration and action in each target. It therefore produces both in a single
generation. Both variants use equal amounts of training data and five random seeds. We evaluate
each trained model on 500 tasks per environment.

\begin{figure}[!htbp]
  \centering
  \begin{minipage}[t]{0.485\linewidth}
    \centering
    \includegraphics[width=\linewidth]{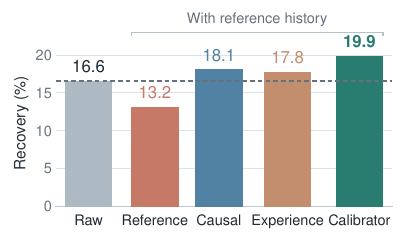}
    \caption{Recovery from the same $L/3$ state (1,000 continuations/condition).
    All but Raw add reference history; the dashed line marks Raw. Experience is generated once at resumption
    and retained; Calibrator runs during continuation.}
    \label{fig:continuation_reference_conditions}
  \end{minipage}\hfill
  \begin{minipage}[t]{0.485\linewidth}
    \centering
    \includegraphics[width=\linewidth,trim=0 3bp 0 3bp,clip]{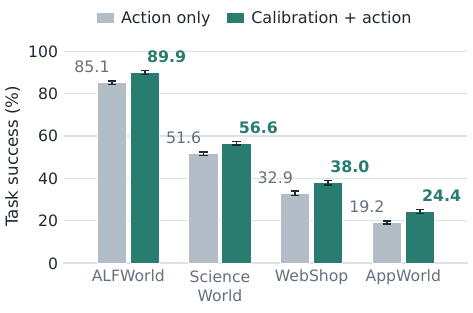}
  \caption{Qwen3-8B actor success. The joint actor generates calibration and action in one response.
    Bars show mean $\pm$ SD across five training seeds, with 500 tasks per environment and seed.}
    \label{fig:joint_calibration_actor_success}
  \end{minipage}
\end{figure}

The joint actor achieves higher task success in all four environments
(Figure~\ref{fig:joint_calibration_actor_success}). Averaged across environments and seeds,
success rises from 47.21\% to 52.21\%. Calibration-augmented distillation therefore improves
task completion performance.

\section{Conclusion}

Interaction history provides evidence for revising earlier attempts, but retaining it does not
ensure better decisions. Labeling observations as action outcomes improves task success without
new environmental information, whereas self-generated experience can introduce redundant or
unsupported guidance. Learning which experiences to retain improves task completion and recovery
when unsuccessful attempts are supplied as reference. Effective history use therefore depends
on using past outcomes to guide later decisions.

\newpage

\FloatBarrier
\subsection*{AI use statement}

We used generative AI tools to assist with literature search, discussion and refinement of
research hypotheses and experimental designs, code implementation and debugging, data analysis,
and figure preparation. We also used these tools to revise, polish, and review the
manuscript. The authors reviewed and verified the AI-assisted work.

\begingroup
\setlength{\emergencystretch}{2em}
\sloppy
\renewcommand{\bibfont}{\small}
\bibliography{references}

\begin{thebibliography}{17}
\providecommand{\natexlab}[1]{#1}
\providecommand{\url}[1]{\texttt{#1}}
\expandafter\ifx\csname urlstyle\endcsname\relax
  \providecommand{\doi}[1]{doi: #1}\else
  \providecommand{\doi}{doi: \begingroup \urlstyle{rm}\Url}\fi

\bibitem[Chen et~al.(2026)Chen, Zhang, Guo, and Zou]{chen2026contextualinertia}
Xingwu Chen, Zhanqiu Zhang, Steven~Y. Guo, and Difan Zou.
\newblock Breaking contextual inertia: Reinforcement learning with single-turn
  anchors for stable multi-turn interaction.
\newblock In \emph{Findings of the Association for Computational Linguistics:
  ACL 2026}, pp.\  6302--6318. Association for Computational Linguistics, 2026.
\newblock \doi{10.18653/v1/2026.findings-acl.313}.
\newblock URL \url{https://aclanthology.org/2026.findings-acl.313/}.

\bibitem[Ding et~al.(2026)Ding, Sun, Zhang, Zhang, and Liu]{ding2026apex}
Jie Ding, Rui Sun, Xinyuan Zhang, Zeyu Zhang, and Xin Liu.
\newblock {APEx}: Distillation of agent procedural experience for adaptive deep
  research question answering.
\newblock \emph{arXiv preprint arXiv:2609.02253}, 2026.
\newblock \doi{10.48550/arXiv.2609.02253}.
\newblock URL \url{https://arxiv.org/abs/2609.02253}.

\bibitem[Feng et~al.(2025)Feng, Xue, Liu, and An]{GiGPO}
Lang Feng, Zhenghai Xue, Tingcong Liu, and Bo~An.
\newblock Group-in-group policy optimization for {LLM} agent training.
\newblock In \emph{Advances in Neural Information Processing Systems},
  volume~38, 2025.
\newblock \doi{10.52202/085713-1544}.
\newblock URL
  \url{https://proceedings.neurips.cc/paper_files/paper/2025/hash/420c9f777c0b4f78d515e53cf74d58b2-Abstract-Conference.html}.

\bibitem[Gumaan(2026)]{gumaan2026feedback}
Esmail Gumaan.
\newblock Feedback that backfires: Why small language model agents repeat the
  call they just watched fail.
\newblock \emph{arXiv preprint arXiv:2608.23651}, 2026.
\newblock \doi{10.48550/arXiv.2608.23651}.
\newblock URL \url{https://arxiv.org/abs/2608.23651}.

\bibitem[Li et~al.(2025)Li, Zhao, Zhou, and Hu]{li2025dora}
Kun Li, Tingzhang Zhao, Wei Zhou, and Songlin Hu.
\newblock {DORA}: Dynamic optimization prompt for continuous reflection of
  {LLM}-based agent.
\newblock In \emph{Proceedings of the 31st International Conference on
  Computational Linguistics}, pp.\  7546--7557. Association for Computational
  Linguistics, 2025.
\newblock URL \url{https://aclanthology.org/2025.coling-main.504/}.

\bibitem[Lippmann et~al.(2025)Lippmann, Spaan, and Yang]{lippmann2025positive}
Philip Lippmann, Matthijs T.~J. Spaan, and Jie Yang.
\newblock Positive experience reflection for agents in interactive text
  environments.
\newblock In \emph{Proceedings of the 1st Workshop for Research on Agent
  Language Models (REALM 2025)}, pp.\  131--142. Association for Computational
  Linguistics, 2025.
\newblock \doi{10.18653/v1/2025.realm-1.10}.
\newblock URL \url{https://aclanthology.org/2025.realm-1.10/}.

\bibitem[Shinn et~al.(2023)Shinn, Cassano, Gopinath, Narasimhan, and
  Yao]{shinn2023reflexion}
Noah Shinn, Federico Cassano, Ashwin Gopinath, Karthik Narasimhan, and Shunyu
  Yao.
\newblock Reflexion: Language agents with verbal reinforcement learning.
\newblock In \emph{Advances in Neural Information Processing Systems},
  volume~36, pp.\  8634--8652, 2023.
\newblock \doi{10.52202/075280-0377}.
\newblock URL \url{https://doi.org/10.52202/075280-0377}.

\bibitem[Shridhar et~al.(2021)Shridhar, Yuan, C{\^o}t{\'e}, Bisk, Trischler,
  and Hausknecht]{shridhar2021alfworld}
Mohit Shridhar, Xingdi Yuan, Marc-Alexandre C{\^o}t{\'e}, Yonatan Bisk, Adam
  Trischler, and Matthew Hausknecht.
\newblock {ALFWorld}: Aligning text and embodied environments for interactive
  learning.
\newblock In \emph{International Conference on Learning Representations}, 2021.
\newblock URL \url{https://openreview.net/forum?id=0IOX0YcCdTn}.

\bibitem[Trivedi et~al.(2024)Trivedi, Khot, Hartmann, Manku, Dong, Li, Gupta,
  Sabharwal, and Balasubramanian]{trivedi2024appworld}
Harsh Trivedi, Tushar Khot, Mareike Hartmann, Ruskin Manku, Vinty Dong, Edward
  Li, Shashank Gupta, Ashish Sabharwal, and Niranjan Balasubramanian.
\newblock {AppWorld}: A controllable world of apps and people for benchmarking
  interactive coding agents.
\newblock In \emph{Proceedings of the 62nd Annual Meeting of the Association
  for Computational Linguistics (Volume 1: Long Papers)}, pp.\  16022--16076,
  2024.
\newblock \doi{10.18653/v1/2024.acl-long.850}.
\newblock URL \url{https://aclanthology.org/2024.acl-long.850/}.

\bibitem[Wan et~al.(2026)Wan, Cao, Zhang, Zeng, Shen, Meng, and
  Zhu]{wan2026inertia}
Yang Wan, Zheng Cao, Zhenhao Zhang, Zhengwen Zeng, Shuheng Shen, Changhua Meng,
  and Linchao Zhu.
\newblock Mitigating conversational inertia in multi-turn agents.
\newblock \emph{arXiv preprint arXiv:2602.03664}, 2026.
\newblock URL \url{https://arxiv.org/abs/2602.03664}.

\bibitem[Wang et~al.(2026)Wang, Leong, Wang, and Li]{wang2026beliefinertia}
Hanlin Wang, Chak~Tou Leong, Jian Wang, and Wenjie Li.
\newblock Seeing isn't believing: Mitigating belief inertia via active
  intervention in embodied agents.
\newblock In \emph{Findings of the Association for Computational Linguistics:
  ACL 2026}, pp.\  37796--37815. Association for Computational Linguistics,
  2026.
\newblock \doi{10.18653/v1/2026.findings-acl.1884}.
\newblock URL \url{https://aclanthology.org/2026.findings-acl.1884/}.

\bibitem[Wang et~al.(2022)Wang, Jansen, C{\^o}t{\'e}, and
  Ammanabrolu]{wang2022scienceworld}
Ruoyao Wang, Peter Jansen, Marc-Alexandre C{\^o}t{\'e}, and Prithviraj
  Ammanabrolu.
\newblock {ScienceWorld}: Is your agent smarter than a 5th grader?
\newblock In \emph{Proceedings of the 2022 Conference on Empirical Methods in
  Natural Language Processing}, pp.\  11279--11298, 2022.
\newblock \doi{10.18653/v1/2022.emnlp-main.775}.
\newblock URL \url{https://aclanthology.org/2022.emnlp-main.775/}.

\bibitem[Xu et~al.(2026)Xu, Yang, and Huang]{xu2026hyperskill}
Ruiyao Xu, Tiankai Yang, and Wei-Chieh Huang.
\newblock {HyperSkill}: Self-evolving {LLM} agents via hypergraph-structured
  skill memory.
\newblock \emph{arXiv preprint arXiv:2608.16114}, 2026.
\newblock \doi{10.48550/arXiv.2608.16114}.
\newblock URL \url{https://arxiv.org/abs/2608.16114}.

\bibitem[Yao et~al.(2022)Yao, Chen, Yang, and Narasimhan]{yao2022webshop}
Shunyu Yao, Howard Chen, John Yang, and Karthik Narasimhan.
\newblock {WebShop}: Towards scalable real-world web interaction with grounded
  language agents.
\newblock In \emph{Advances in Neural Information Processing Systems},
  volume~35, pp.\  20744--20757, 2022.
\newblock \doi{10.52202/068431-1508}.
\newblock URL \url{https://doi.org/10.52202/068431-1508}.

\bibitem[Yao et~al.(2023)Yao, Zhao, Yu, Du, Shafran, Narasimhan, and
  Cao]{yao2023react}
Shunyu Yao, Jeffrey Zhao, Dian Yu, Nan Du, Izhak Shafran, Karthik Narasimhan,
  and Yuan Cao.
\newblock {ReAct}: Synergizing reasoning and acting in language models.
\newblock In \emph{International Conference on Learning Representations}, 2023.
\newblock URL \url{https://openreview.net/forum?id=WE_vluYUL-X}.

\bibitem[Zhang et~al.(2026{\natexlab{a}})Zhang, Long, Bao, Feng, Zhang, Yue,
  and Wang]{zhang2026memskill}
Haozhen Zhang, Quanyu Long, Jianzhu Bao, Tao Feng, Weizhi Zhang, Haodong Yue,
  and Wenya Wang.
\newblock {MemSkill}: Learning and evolving memory skills for self-evolving
  agents.
\newblock \emph{arXiv preprint arXiv:2602.02474}, 2026{\natexlab{a}}.
\newblock \doi{10.48550/arXiv.2602.02474}.
\newblock URL \url{https://arxiv.org/abs/2602.02474}.

\bibitem[Zhang et~al.(2026{\natexlab{b}})Zhang, Wang, Zhou, Liao, Feng, Li,
  Zheng, Zhang, Wen, Li, Xiong, Qi, Tang, and Wen]{zhang2026memrl}
Shengtao Zhang, Jiaqian Wang, Ruiwen Zhou, Junwei Liao, Yuchen Feng, Zhuo Li,
  Yujie Zheng, Weinan Zhang, Ying Wen, Zhiyu Li, Feiyu Xiong, Yutao Qi,
  Bo~Tang, and Muning Wen.
\newblock {MemRL}: Self-evolving agents via runtime reinforcement learning on
  episodic memory.
\newblock \emph{arXiv preprint arXiv:2601.03192}, 2026{\natexlab{b}}.
\newblock \doi{10.48550/arXiv.2601.03192}.
\newblock URL \url{https://arxiv.org/abs/2601.03192}.

\end{thebibliography}
\bibliographystyle{iclr2027_conference}
\endgroup

\clearpage
\appendix

\section{Shared Experimental Protocol}
\label{app:experimental_protocol}

\paragraph{Interaction and termination.}
Full rollouts for the history interventions and calibration comparisons allow at most 30 actions
and terminate upon success, terminal failure, or exhaustion of this budget. Continuations restore
the environment to $L/3$ and receive 30 additional actions. Success is determined by the
environment's final \nolinkurl{won} or equivalent \nolinkurl{success} flag, not intermediate
progress scores. Full-rollout Experience uses separate calibration and action calls, each capped
at 8,192 output tokens.

\Needspace{10\baselineskip}
\section{Fixed-History Audits}
\label{app:motivation_evidence}

\FloatBarrier

\subsection{Next-Action Repetition on Fixed Histories}
\label{app:loop2_same_source_repetition}

Exact repetition can be a valid part of a task. To focus on reuse despite contrary feedback, this
probe reuses a fixed cohort of 1,815 histories ending in two identical actions, selected by
screening for feedback that does not support another execution of the same action (judged by
GPT-5.6-Sol). The source pool consists of 500 natural rollouts per model--environment cell. We
select at most one pair from each trajectory: its earliest nonterminal repeat pair. The same actor
that produced a history receives it under each
condition. This fixes the task, preceding actions, and environmental observations across the
three conditions discussed in Sections~\ref{sec:fixed_history_repetition}
and~\ref{sec:self_persistent_calibration}.

\paragraph{Three renderings of the same prefix.}
Raw is the vanilla interaction history.
Causal preserves this sequence and labels each returned observation in place as the outcome of
its preceding action. Experience first asks the source actor to
review the Raw prefix in a separate calibration call. A generated experience is inserted beside
the latest action--outcome transition before the next decision; \texttt{NO\_UPDATE} or an invalid
update leaves the Raw input unchanged.

\begin{table}[!htbp]
  \centering
  \caption{Next-action repetition on fixed histories. Entries show repeated-action counts with
  rates in parentheses. $N$ is the shared number of histories for each condition.
  Figure~\ref{fig:explicit_interfaces_repetition} plots these rates.}
  \label{tab:loop2_filtered_repetition_marginals}
  \setlength{\tabcolsep}{5pt}
  \renewcommand{\arraystretch}{1.1}
  \begin{tabular}{@{}lrccc@{}}
\toprule
\textbf{Model} & $N$ & \textbf{Raw} & \textbf{Causal} & \textbf{Experience} \\
\midrule
GPT-5.6-Sol & 46 & 4 (8.70\%) & 3 (6.52\%) & 2 (4.35\%) \\
DeepSeek-V4-Pro & 131 & 33 (25.19\%) & 30 (22.90\%) & 25 (19.08\%) \\
Qwen3.5-Plus & 188 & 48 (25.53\%) & 41 (21.81\%) & 36 (19.15\%) \\
Qwen3-32B & 492 & 93 (18.90\%) & 77 (15.65\%) & 84 (17.07\%) \\
Qwen3-8B & 958 & 492 (51.36\%) & 434 (45.30\%) & 241 (25.16\%) \\
\midrule
\textbf{Pooled} & \textbf{1,815} & \textbf{670 (36.91\%)} & \textbf{585 (32.23\%)} & \textbf{388 (21.38\%)} \\
\bottomrule
\end{tabular}

\end{table}
\FloatBarrier

\section{Complete Diagnostic Success Results}
\label{app:action_history_full}
\label{app:action_history_cells}
\label{app:explicit_outcome_success}

Table~\ref{tab:action_history_cells} combines the history interventions and explicit-calibration
diagnostics from Sections~\ref{sec:action_history_utilization} and~\ref{sec:explicit_calibration}.
All seven conditions use the same 500 tasks per model--environment cell and share the Raw baseline.

\begin{table}[!htbp]
  \centering
  \caption{Complete diagnostic success counts, out of 500 tasks per model--environment cell.
  Totals aggregate 10,000 episodes per condition. ``Obs.-Only'' denotes Observation-Only;
  ``Exp.'' denotes Experience.}
  \label{tab:action_history_cells}
  \label{tab:explicit_outcome_cells}
  \setlength{\tabcolsep}{2.2pt}
  \renewcommand{\arraystretch}{1.08}
\begin{tabular}{@{}llrrrrrrr@{}}
\toprule
\textbf{Model} & \textbf{Environment} & \textbf{Raw} & \textbf{\shortstack{Recent-3-\\Only}} & \textbf{\shortstack{Obs.-\\Only}} & \textbf{\shortstack{Shuffle-\\All}} & \textbf{\shortstack{Shuffle-\\Old}} & \textbf{Causal} & \textbf{Exp.} \\
\midrule
\multirow{4}{*}{GPT-5.6-Sol} & ALFWorld & 448 & 421 & 456 & 456 & 453 & 449 & 441 \\
 & ScienceWorld & 403 & 338 & 384 & 364 & 400 & 410 & 394 \\
 & WebShop & 188 & 171 & 143 & 169 & 186 & 190 & 176 \\
 & AppWorld & 427 & 274 & 394 & 426 & 430 & 437 & 421 \\
\addlinespace[0.35em]
\multirow{4}{*}{DeepSeek-V4-Pro} & ALFWorld & 350 & 280 & 318 & 327 & 349 & 357 & 346 \\
 & ScienceWorld & 293 & 235 & 267 & 255 & 285 & 297 & 282 \\
 & WebShop & 185 & 168 & 109 & 146 & 180 & 187 & 174 \\
 & AppWorld & 433 & 245 & 426 & 425 & 433 & 435 & 431 \\
\addlinespace[0.35em]
\multirow{4}{*}{Qwen3.5-Plus} & ALFWorld & 327 & 255 & 349 & 374 & 362 & 355 & 358 \\
 & ScienceWorld & 169 & 134 & 136 & 139 & 169 & 195 & 171 \\
 & WebShop & 163 & 177 & 110 & 161 & 166 & 176 & 169 \\
 & AppWorld & 307 & 177 & 227 & 286 & 299 & 313 & 288 \\
\addlinespace[0.35em]
\multirow{4}{*}{Qwen3-32B} & ALFWorld & 308 & 196 & 223 & 270 & 302 & 314 & 269 \\
 & ScienceWorld & 75 & 64 & 73 & 74 & 75 & 101 & 86 \\
 & WebShop & 92 & 98 & 78 & 91 & 93 & 98 & 80 \\
 & AppWorld & 106 & 101 & 72 & 109 & 126 & 111 & 101 \\
\addlinespace[0.35em]
\multirow{4}{*}{Qwen3-8B} & ALFWorld & 244 & 159 & 201 & 189 & 236 & 244 & 183 \\
 & ScienceWorld & 38 & 26 & 37 & 39 & 35 & 50 & 48 \\
 & WebShop & 65 & 68 & 77 & 69 & 72 & 84 & 66 \\
 & AppWorld & 18 & 12 & 13 & 15 & 24 & 22 & 17 \\
\midrule
\multicolumn{2}{l}{Total successes / 10,000} & 4,639 & 3,599 & 4,093 & 4,384 & 4,675 & 4,825 & 4,501 \\
\multicolumn{2}{l}{Pooled success rate} & 46.39\% & 35.99\% & 40.93\% & 43.84\% & 46.75\% & 48.25\% & 45.01\% \\
\bottomrule
\end{tabular}

\end{table}

\FloatBarrier
\section{Causal Transition and Experience Diagnostics}
\label{app:calibration_repetition_full}

\subsection{Explicit History Rechecking}
\label{app:history_check}

Check retains the Raw history format and adds the following reminder to the actor prompt:
\begin{quote}
\raggedright
\texttt{[History caution]}\par
Earlier actions, reasoning, plans, and interpretations in the interaction history may be wrong,
incomplete, or stale. Re-check them against the task goal, constraints, and recorded environment
or tool observations before choosing the next action. Repetition is not evidence of correctness.\par
\texttt{[/History caution]}
\end{quote}
Table~\ref{tab:history_check_success} reports the full model--environment results.
Overall, Check improves on Raw and achieves a similar success rate to Causal Transition.

\begin{table}[!htbp]
  \centering
  \caption{Task success with an explicit history-check reminder. Entries are successful-task
  counts out of 500 per model--environment cell. Overall totals pool 10,000 tasks per condition;
  the final row gives the corresponding success rates. Causal denotes Causal Transition.}
  \label{tab:history_check_success}
  \setlength{\tabcolsep}{9pt}
  \renewcommand{\arraystretch}{1.08}
  \begin{tabular}{@{}llrrr@{}}
    \toprule
    \textbf{Model} & \textbf{Environment} & \textbf{Raw} & \textbf{Causal} & \textbf{Check} \\
    \midrule
    \multirow{4}{*}{GPT-5.6-Sol} & ALFWorld & 448 & 449 & 467 \\
     & ScienceWorld & 403 & 410 & 410 \\
     & WebShop & 188 & 190 & 195 \\
     & AppWorld & 427 & 437 & 417 \\
    \addlinespace[0.35em]
    \multirow{4}{*}{DeepSeek-V4-Pro} & ALFWorld & 350 & 357 & 359 \\
     & ScienceWorld & 293 & 297 & 305 \\
     & WebShop & 185 & 187 & 190 \\
     & AppWorld & 433 & 435 & 425 \\
    \addlinespace[0.35em]
    \multirow{4}{*}{Qwen3.5-Plus} & ALFWorld & 327 & 355 & 362 \\
     & ScienceWorld & 169 & 195 & 200 \\
     & WebShop & 163 & 176 & 179 \\
     & AppWorld & 307 & 313 & 290 \\
    \addlinespace[0.35em]
    \multirow{4}{*}{Qwen3-32B} & ALFWorld & 308 & 314 & 315 \\
     & ScienceWorld & 75 & 101 & 107 \\
     & WebShop & 92 & 98 & 95 \\
     & AppWorld & 106 & 111 & 117 \\
    \addlinespace[0.35em]
    \multirow{4}{*}{Qwen3-8B} & ALFWorld & 244 & 244 & 267 \\
     & ScienceWorld & 38 & 50 & 52 \\
     & WebShop & 65 & 84 & 85 \\
     & AppWorld & 18 & 22 & 23 \\
    \midrule
    \multicolumn{2}{l}{Overall successes} & 4,639 & 4,825 & 4,860 \\
    \multicolumn{2}{l}{Overall success (\%)} & 46.39 & 48.25 & 48.60 \\
    \bottomrule
  \end{tabular}
\end{table}

\subsection{Retaining Experience in Later Prompts}
\label{app:calibration_interface_examples}
\label{app:self_persistent_protocol}

After each nonterminal transition, a separate call to the same actor model receives the environment, task,
preceding raw history, latest action, and returned observation. Earlier generated experiences,
calibration reasoning, rewards, success labels, and future actions are excluded. Its final JSON is
one of
\begin{quote}
\texttt{\{"decision":"NO\_UPDATE","experience":null\}}\par
\texttt{\{"decision":"CALIBRATE","experience":"..."\}}.
\end{quote}
The prompt asks for a new task-relevant correction or durable progress fact that should affect
later decisions. There is no calibration before the initial action; terminal transitions are
recorded and skipped. Each generated experience is linked to its source transition and rendered
beside it in every subsequent action prompt.

\subsection{Experience Generation Rates}
\label{app:experience_decision_rates}

We measure how often actors write an experience in an additional set of full rollouts under
the same Experience protocol, with 500 tasks per model--environment cell.
Table~\ref{tab:experience_decision_rates} reports the numbers of generated experiences and
\texttt{NO\_UPDATE} responses, together with their percentages among valid review responses.

\begin{table}[!htbp]
  \centering
  \caption{Experience generation and \texttt{NO\_UPDATE} decisions across five actors and four
  environments. Entries give counts with percentages of valid review responses in parentheses.
  Each cell covers 500 tasks; Overall pools responses across all 10,000 episodes.}
  \label{tab:experience_decision_rates}
  \setlength{\tabcolsep}{5pt}
  \renewcommand{\arraystretch}{1.08}
\begin{tabular}{@{}llrrr@{}}
\toprule
\textbf{Model} & \textbf{Environment} & \textbf{Valid reviews} & \textbf{Experience} & \textbf{NO\_UPDATE} \\
\midrule
\multirow{4}{*}{GPT-5.6-Sol} & ALFWorld & 5,848 & 4,182 (71.51) & 1,666 (28.49) \\
 & ScienceWorld & 7,044 & 4,723 (67.05) & 2,321 (32.95) \\
 & WebShop & 3,453 & 1,886 (54.62) & 1,567 (45.38) \\
 & AppWorld & 5,636 & 5,482 (97.27) & 154 (2.73) \\
\addlinespace[0.35em]
\multirow{4}{*}{DeepSeek-V4-Pro} & ALFWorld & 8,605 & 1,980 (23.01) & 6,625 (76.99) \\
 & ScienceWorld & 8,757 & 2,471 (28.22) & 6,286 (71.78) \\
 & WebShop & 3,535 & 1,385 (39.18) & 2,150 (60.82) \\
 & AppWorld & 4,899 & 2,326 (47.48) & 2,573 (52.52) \\
\addlinespace[0.35em]
\multirow{4}{*}{Qwen3.5-Plus} & ALFWorld & 8,117 & 861 (10.61) & 7,256 (89.39) \\
 & ScienceWorld & 8,990 & 1,167 (12.98) & 7,823 (87.02) \\
 & WebShop & 3,288 & 876 (26.64) & 2,412 (73.36) \\
 & AppWorld & 5,168 & 2,182 (42.22) & 2,986 (57.78) \\
\addlinespace[0.35em]
\multirow{4}{*}{Qwen3-32B} & ALFWorld & 10,241 & 1,338 (13.07) & 8,903 (86.93) \\
 & ScienceWorld & 8,845 & 1,223 (13.83) & 7,622 (86.17) \\
 & WebShop & 2,199 & 643 (29.24) & 1,556 (70.76) \\
 & AppWorld & 5,029 & 2,744 (54.56) & 2,285 (45.44) \\
\addlinespace[0.35em]
\multirow{4}{*}{Qwen3-8B} & ALFWorld & 11,546 & 6,851 (59.34) & 4,695 (40.66) \\
 & ScienceWorld & 10,508 & 6,961 (66.24) & 3,547 (33.76) \\
 & WebShop & 2,249 & 2,155 (95.82) & 94 (4.18) \\
 & AppWorld & 7,634 & 6,053 (79.29) & 1,581 (20.71) \\
\midrule
\multicolumn{2}{l}{Overall} & 131,591 & 57,489 (43.69) & 74,102 (56.31) \\
\bottomrule
\end{tabular}

\end{table}
\FloatBarrier

\subsection{Candidate Experience and Next-Action Quality}
\label{app:experience_next_action_quality}

This comparison examines whether a generated experience helps an actor choose
a better next action from the same interaction history.

GPT-5.6-Sol generates candidate experiences from real rollout contexts. For each context,
each of the five recipient actors generates two next actions: one with the candidate experience
and one without it. These actions form a within-actor pair for evaluation.

\paragraph{Pairwise evaluation.}
When the parsed executable actions are identical, the pair is assigned Tie without judging;
reasoning text need not match. Other pairs are evaluated by GPT-5.6-Sol in two blind comparisons
with the A/B order reversed, at temperature 0. Win denotes better next-action quality with
the experience; Loss denotes worse quality. These outcomes require agreement across both
orders. Two Tie judgments or inconsistent judgments are recorded as Tie.

\begin{figure}[!htbp]
  \centering
  \includegraphics[width=\linewidth]{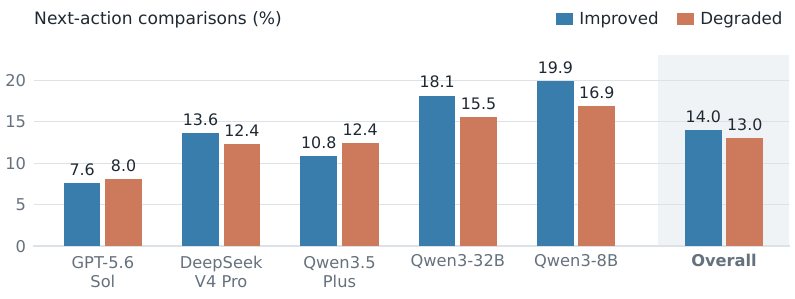}
  \caption{LLM-judged changes in next-action quality with candidate experiences generated by
  GPT-5.6-Sol. Bars compare actions generated by each recipient actor with and without an experience
  on the same history. Rates use all completed comparisons, including ties; ties are omitted from the bars.}
  \label{fig:experience_next_action_quality}
\end{figure}

\paragraph{Effect on next-action quality.}
Across all completed comparisons, the paired judgments favor the action generated with an
experience in 14.0\% of cases and favor the Raw action in 13.0\%
(Figure~\ref{fig:experience_next_action_quality}). All five recipient actors show both outcomes.
Under this evaluation, helpful guidance coexists with guidance that leads to worse next
actions, even though all candidates are generated by GPT-5.6-Sol. Explicit experience generation
therefore does not reliably yield an improvement in next-action quality.

\subsection{Generated Experiences and Their Subsequent Use}
\label{app:self_persistent_cases}

Table~\ref{tab:self_persistent_cases} contrasts a redundant restatement, an unsupported claim,
and an experience that calls for further verification, drawn from additional Experience rollouts.
We verify that each experience appears in the next actor prompt alongside its source transition;
steps are zero-based.
Figure~\ref{fig:experience_quality_cases} presents two additional ALFWorld examples from
GPT-5.6-Sol and Qwen3.5-Plus, contrasting a restated outcome with an unsupported operational rule.

\begin{table}[!htbp]
  \centering
  \caption{Examples of retained judgments and subsequent actions, with verbatim experience excerpts.}
  \label{tab:self_persistent_cases}
  \setlength{\tabcolsep}{4pt}
  \renewcommand{\arraystretch}{1.10}
  \begin{tabularx}{\linewidth}{@{}>{\raggedright\arraybackslash}p{0.14\linewidth}>{\raggedright\arraybackslash}p{0.49\linewidth}>{\raggedright\arraybackslash}X@{}}
    \toprule
    \textbf{Case} & \textbf{Task evidence and retained experience} & \textbf{Subsequent behavior} \\
    \midrule
    \textbf{Restatement}\newline ALFWorld\newline Qwen3-8B &
    The task requires cleaning an egg and placing it in a fridge.
    \texttt{put egg 2 in bowl 1} returns ``Nothing happens.''\par\smallskip
    \textit{Experience:} ``Putting the egg in a bowl does not successfully place it in the fridge.'' &
    The action and identical experience recur at steps 11--13, with earlier judgments visible.
    Repeated writes restate an already recorded failure. \\
    \addlinespace[0.5em]
    \textbf{Unsupported}\newline\textbf{claim}\newline WebShop\newline Qwen3.5-Plus &
    A 3.4-oz, paraben-free shampoo below \$40 is requested. The opened page shows a 3.5-oz
    product at \$29.99 but does not verify the paraben-free attribute.\par\smallskip
    \textit{Experience:} ``The selected 3.5 oz dry shampoo is paraben-free and priced under \$40,
    confirming that product detail pages provide the necessary attribute verification to satisfy
    the user's specific constraints.'' &
    The actor buys the product next without inspecting features or description.
    The judgment overlooks the size mismatch and endorses an unverified attribute. \\
    \addlinespace[0.5em]
    \textbf{Verification}\newline WebShop\newline Qwen3.5-Plus &
    A black, non-slip PU leather organizer below \$80 is requested. The page verifies color,
    material, and a \$45.99 price, but not non-slip functionality.\par\smallskip
    \textit{Experience excerpt:} ``\ldots{} though non-slip functionality requires further
    verification via features or description.'' &
    The actor selects black, then inspects features at step 3 and description at step 5.
    The experience notes that the non-slip property still needs verification. \\
    \bottomrule
  \end{tabularx}
\end{table}

\FloatBarrier
\subsection{Reference-History Recovery from a Shared State}
\label{app:reference_recovery}
\label{app:failure_conditioned_continuation}
\label{app:one_shot_recovery}
\label{sec:one_shot_recovery}

\paragraph{Shared restored state and history conditions.}
All five conditions use the 1,000 unsuccessful source trajectories selected in
Section~\ref{sec:motivation}, with 50 sources per model--environment cell. Every continuation
restores the environment to $\lfloor L/3\rfloor$ and uses the source actor with a budget of
30 new actions after restoration. Raw supplies only the history through this point. Reference
adds the recorded actions and observations between $\lfloor L/3\rfloor$ and
$\lfloor2L/3\rfloor$ as reference information, without executing them in the restored
environment. The three intervention conditions use the same reference segment, giving
1,000 continuations per condition and 5,000 in total.

\paragraph{Separating reference history from actual interaction.}
The additional segment is placed in a delimited reference block. The prompt then distinguishes
newly executed continuation history from that reference and supplies the actual current
observation and available actions. At the first continuation step, the newly executed
history is \texttt{None}. The reference-augmented prompt has the following structure:
\begin{quote}
\begin{minipage}{\linewidth}
\ttfamily\raggedright
\{Normal history through L/3\}\par\medskip
Additional reference history\par
(not executed in the current environment; the actual environment was restored to L/3):\par
\{Reference history from L/3 to 2L/3\}\par\medskip
End of additional reference history.\par\medskip
Actual interaction history after restoring the environment to L/3:\par
\{Actual continuation history; None at the first step\}\par\medskip
Current observation:\par
\{Actual current observation\}\par\medskip
Available actions:\par
\{Actual current actions\}
\end{minipage}
\end{quote}

\paragraph{Feedback organization and calibration.}
Reference retains the ordinary observation--action format. Causal + Reference instead
labels each returned observation as the outcome of its preceding action throughout the actual history, reference
segment, and newly executed continuation. Reference records are explicitly labeled as
\texttt{Reference Action} and \texttt{Outcome of Reference Action}; their boundaries remain
separate from the actual interaction.
Experience retains the ordinary history format, generates an experience,
and keeps that experience available for subsequent decisions. Calibrator applies the frozen
model and update rule from Section~\ref{sec:outcome_triggered_calibration} to the
reference-augmented continuation, retaining generated updates while the source actor
continues to choose actions.

\FloatBarrier
\section{Additional Evaluation of the Calibrator}
\label{app:calibrator_method_and_evaluation}

\subsection{Full Task-Success Results: 500 Tasks per Cell}
\label{app:explicit_calibration_success_cells}

Table~\ref{tab:explicit_calibration_success_full} reports Raw, Causal Transition, and the
trained calibrator with 500 tasks per cell. Entries report task-success rates (\%);
Figure~\ref{fig:outcome_triggered_calibration_success} gives the model and environment marginals.

\begin{table}[!htbp]
  \centering
  \caption{Method-evaluation success rates (\%), with 500 tasks
  per cell. The three conditions each contain 10,000 episodes.}
  \label{tab:explicit_calibration_success_full}
  \setlength{\tabcolsep}{4pt}
  \renewcommand{\arraystretch}{1.0}
\begin{tabular}{@{}llrrrr@{}}
\toprule
\textbf{Model} & \textbf{Setting} & \textbf{ALFWorld} & \textbf{ScienceWorld} & \textbf{WebShop} & \textbf{AppWorld} \\
\midrule
\multirow{3}{*}{GPT-5.6-Sol} & Raw & 89.60 & 80.60 & 37.60 & 85.40 \\
 & Causal Transition & 89.80 & 82.00 & 38.00 & 87.40 \\
 & Calibrator & 94.20 & 91.80 & 42.40 & 87.00 \\
\addlinespace[0.35em]
\multirow{3}{*}{DeepSeek-V4-Pro} & Raw & 70.00 & 58.60 & 37.00 & 86.60 \\
 & Causal Transition & 71.40 & 59.40 & 37.40 & 87.00 \\
 & Calibrator & 75.00 & 67.40 & 43.40 & 84.20 \\
\addlinespace[0.35em]
\multirow{3}{*}{Qwen3.5-Plus} & Raw & 65.40 & 33.80 & 32.60 & 61.40 \\
 & Causal Transition & 71.00 & 39.00 & 35.20 & 62.60 \\
 & Calibrator & 80.40 & 37.80 & 38.60 & 61.00 \\
\addlinespace[0.35em]
\multirow{3}{*}{Qwen3-32B} & Raw & 61.60 & 15.00 & 18.40 & 21.20 \\
 & Causal Transition & 62.80 & 20.20 & 19.60 & 22.20 \\
 & Calibrator & 60.80 & 22.00 & 23.00 & 23.20 \\
\addlinespace[0.35em]
\multirow{3}{*}{Qwen3-8B} & Raw & 48.80 & 7.60 & 13.00 & 3.60 \\
 & Causal Transition & 48.80 & 10.00 & 16.80 & 4.40 \\
 & Calibrator & 51.20 & 13.20 & 23.40 & 9.00 \\
\bottomrule
\end{tabular}

\end{table}

\FloatBarrier
\subsection{Ablating Paired Next-Action Comparisons}
\label{app:calibrator_content_ablation}

The \emph{content-screening-only} ablation uses GPT's content screening of candidate experiences
for unsupported claims and redundant restatements of observations to construct training targets.
It omits the subsequent generation and comparison of next actions with and
without experience for each actor. Table~\ref{tab:calibrator_content_ablation}
compares the two trained calibrators on task success, with 500 tasks per model--environment cell.
Pooled success increases from 49.61\% with content screening alone to 51.45\% with the full
pipeline, with gains in every model-level and environment-level aggregate.

\begin{table}[!htbp]
  \centering
  \caption{Training-data ablation of paired next-action comparisons. Entries are successful-task
  counts out of 500 per model--environment cell. Content screening only retains GPT's review of
  the experience text but omits the subsequent action comparisons; Full Calibrator uses both
  stages. The final two rows pool 10,000 tasks per condition.}
  \label{tab:calibrator_content_ablation}
  \setlength{\tabcolsep}{2.5pt}
  \renewcommand{\arraystretch}{1.08}
\begin{tabular}{@{}llrrrr@{}}
\toprule
\textbf{Model} & \textbf{Training data} & \textbf{ALFWorld} & \textbf{ScienceWorld} & \textbf{WebShop} & \textbf{AppWorld} \\
\midrule
\multirow{2}{*}{GPT-5.6-Sol} & Content screening only & 452 & 451 & 204 & 424 \\
 & Full Calibrator & 471 & 459 & 212 & 435 \\
\addlinespace[0.35em]
\multirow{2}{*}{DeepSeek-V4-Pro} & Content screening only & 364 & 312 & 205 & 417 \\
 & Full Calibrator & 375 & 337 & 217 & 421 \\
\addlinespace[0.35em]
\multirow{2}{*}{Qwen3.5-Plus} & Content screening only & 397 & 197 & 182 & 293 \\
 & Full Calibrator & 402 & 189 & 193 & 305 \\
\addlinespace[0.35em]
\multirow{2}{*}{Qwen3-32B} & Content screening only & 284 & 105 & 107 & 108 \\
 & Full Calibrator & 304 & 110 & 115 & 116 \\
\addlinespace[0.35em]
\multirow{2}{*}{Qwen3-8B} & Content screening only & 252 & 67 & 111 & 29 \\
 & Full Calibrator & 256 & 66 & 117 & 45 \\
\midrule
\multirow{2}{*}{Overall} & Content screening only & \multicolumn{4}{r}{4,961 / 10,000 (49.61\%)} \\
 & Full Calibrator & \multicolumn{4}{r}{5,145 / 10,000 (51.45\%)} \\
\bottomrule
\end{tabular}

\end{table}
\FloatBarrier

\section{History as Separate User and Assistant Messages}
\label{app:native_chat_robustness}

The main experiments present accumulated interactions in a single history block. Here we test
the history interventions and same-state continuation using separate chat messages.
Fixed instructions remain in the system message;
historical observations and executed actions alternate as user and assistant messages. The final
user message contains the current observation and any currently available actions. Historical
available-action lists are not replayed.

\subsection{Full-Rollout History Interventions}

Rollouts allow at most 30 actions, with an explicit
8,192-token output limit per actor request. The seven conditions use the same 100 tasks in each
model--environment cell, totaling 2,000 episodes per condition across five actors and four environments.

The history interventions follow Section~\ref{sec:action_history_utilization}, applied to the
corresponding user and assistant messages. Under Causal Transition, returned observations are
labeled in their existing user messages; Experience uses the separate review calls and retains
generated experiences as described in Appendix~\ref{app:self_persistent_protocol}.

\begin{table}[!htbp]
  \centering
  \caption{Native-chat success counts out of 100 per model--environment cell (2,000 episodes per
  condition). ``Obs. Only'' denotes Observation-Only; ``Exp.'' denotes Experience.}
  \label{tab:native_chat_robustness_cells}
  \setlength{\tabcolsep}{2.75pt}
  \renewcommand{\arraystretch}{1.08}
  \begin{tabular}{@{}llrrrrrrr@{}}
\toprule
\textbf{Model} & \textbf{Environment} & \textbf{Raw} & \shortstack{\textbf{Recent-3}\\\textbf{Only}} & \shortstack{\textbf{Obs.}\\\textbf{Only}} & \shortstack{\textbf{Shuffle}\\\textbf{All}} & \shortstack{\textbf{Shuffle}\\\textbf{Old}} & \textbf{Causal} & \textbf{Exp.} \\
\midrule
\multirow{4}{*}{GPT-5.6-Sol} & ALFWorld & 95 & 80 & 93 & 93 & 95 & 95 & 90 \\
 & ScienceWorld & 83 & 63 & 72 & 79 & 79 & 85 & 80 \\
 & WebShop & 45 & 41 & 31 & 41 & 39 & 45 & 45 \\
 & AppWorld & 81 & 40 & 73 & 80 & 83 & 84 & 82 \\
\addlinespace[0.35em]
\multirow{4}{*}{DeepSeek-V4-Pro} & ALFWorld & 67 & 59 & 66 & 68 & 61 & 65 & 67 \\
 & ScienceWorld & 44 & 27 & 48 & 45 & 52 & 46 & 51 \\
 & WebShop & 34 & 39 & 24 & 26 & 39 & 31 & 31 \\
 & AppWorld & 76 & 44 & 76 & 77 & 75 & 79 & 75 \\
\addlinespace[0.35em]
\multirow{4}{*}{Qwen3.5-Plus} & ALFWorld & 85 & 74 & 79 & 81 & 83 & 85 & 81 \\
 & ScienceWorld & 42 & 22 & 34 & 37 & 36 & 43 & 38 \\
 & WebShop & 33 & 36 & 27 & 33 & 38 & 38 & 37 \\
 & AppWorld & 52 & 29 & 42 & 45 & 55 & 58 & 50 \\
\addlinespace[0.35em]
\multirow{4}{*}{Qwen3-32B} & ALFWorld & 55 & 42 & 49 & 51 & 53 & 57 & 52 \\
 & ScienceWorld & 20 & 15 & 17 & 19 & 17 & 21 & 22 \\
 & WebShop & 19 & 21 & 17 & 19 & 20 & 21 & 17 \\
 & AppWorld & 19 & 18 & 11 & 12 & 15 & 20 & 17 \\
\addlinespace[0.35em]
\multirow{4}{*}{Qwen3-8B} & ALFWorld & 48 & 36 & 42 & 42 & 45 & 50 & 45 \\
 & ScienceWorld & 6 & 2 & 3 & 5 & 3 & 9 & 7 \\
 & WebShop & 13 & 15 & 12 & 12 & 12 & 18 & 14 \\
 & AppWorld & 3 & 2 & 3 & 5 & 11 & 5 & 5 \\
\midrule
\multicolumn{2}{l}{Total successes / 2,000} & 920 & 705 & 819 & 870 & 911 & 955 & 906 \\
\multicolumn{2}{l}{Pooled success rate} & 46.00\% & 35.25\% & 40.95\% & 43.50\% & 45.55\% & 47.75\% & 45.30\% \\
\bottomrule
\end{tabular}

\end{table}
\FloatBarrier

The pooled results retain the main ordering: Causal improves over Raw, whereas Experience
does not. Retaining earlier content remains useful, and Shuffle-Old stays close to Raw while
outperforming Recent-3-Only.

\subsection{Reference-History Continuation}
\label{app:native_chat_continuation}

To test whether the recovery decline in Section~\ref{sec:motivation} also occurs with this
message format, we compare Raw and Reference on 50 paired unsuccessful sources per
model--environment cell, giving 1,000 continuations per condition. Both conditions restore
the environment to $\lfloor L/3\rfloor$ and allow up to 30 new actions by the source actor.
Raw supplies only the actual prefix. Reference adds the recorded segment from
$\lfloor L/3\rfloor$ to $\lfloor2L/3\rfloor$ as a separate user message after that prefix,
explicitly marked as not executed in the restored environment. A subsequent user message
provides the actual current observation and available actions. Newly executed interactions
continue as user and assistant messages.

Reference lowers pooled recovery from 158/1,000 (15.8\%) to 114/1,000 (11.4\%)
(Table~\ref{tab:native_chat_continuation_cells}). Recovery declines in all five model
aggregates and all four environment aggregates. The decline therefore also appears when
actual interaction history is represented as separate chat messages.

\begin{table}[!htbp]
  \centering
  \caption{Recovery success counts with user--assistant history. Each model--environment
  cell contains 50 paired sources; the Total column is out of 200 per model.
  Overall counts are out of 250 per environment and 1,000 in total.}
  \label{tab:native_chat_continuation_cells}
  \begin{tabular}{@{}llrrrrr@{}}
\toprule
Model & History & ALFWorld & ScienceWorld & WebShop & AppWorld & Total \\
\midrule
\multirow{2}{*}{GPT-5.6-Sol} & Raw & 25 & 20 & 4 & 5 & 54 \\
 & Reference & 24 & 17 & 4 & 3 & 48 \\
\addlinespace
\multirow{2}{*}{DeepSeek-V4-Pro} & Raw & 10 & 12 & 9 & 5 & 36 \\
 & Reference & 6 & 9 & 5 & 3 & 23 \\
\addlinespace
\multirow{2}{*}{Qwen3.5-Plus} & Raw & 11 & 12 & 9 & 5 & 37 \\
 & Reference & 5 & 8 & 4 & 4 & 21 \\
\addlinespace
\multirow{2}{*}{Qwen3-32B} & Raw & 10 & 3 & 2 & 4 & 19 \\
 & Reference & 7 & 4 & 3 & 0 & 14 \\
\addlinespace
\multirow{2}{*}{Qwen3-8B} & Raw & 9 & 0 & 1 & 2 & 12 \\
 & Reference & 6 & 0 & 2 & 0 & 8 \\
\midrule
\multirow{2}{*}{Overall} & Raw & 65 & 47 & 25 & 21 & 158 \\
 & Reference & 48 & 38 & 18 & 10 & 114 \\
\bottomrule
\end{tabular}

\end{table}
\FloatBarrier


\newenvironment{coreprompt}{%
  \begin{list}{}{%
    \setlength{\leftmargin}{1em}%
    \setlength{\rightmargin}{1em}%
  }\item[]\begin{minipage}{\linewidth}
  \ttfamily\raggedright\sloppy
}{%
  \end{minipage}\end{list}
}

\section{Core Prompt Interfaces}
\label{app:prompt_templates}

This section records the exact text that defines the paper's central history and
Experience interventions.

\subsection{History Rendering}
\label{app:history_rendering_prompts}

The following abstract templates match the single history block used in the main
evaluation. Raw preserves the ordinary observation--action sequence. Causal Transition
changes only the label that assigns the returned observation to its preceding action.
Experience writes an actor-generated review beside its source transition and retains it in later
actor prompts; \texttt{NO\_UPDATE} inserts no text.

\Needspace{6\baselineskip}
\paragraph{Raw.}
\begin{coreprompt}
[Observation \{t\}: '\{O\_t\}', Action \{t\}: '\{A\_t\}']

[Observation \{t+1\}: '\{O\_\{t+1\}\}', Action \{t+1\}: '\{A\_\{t+1\}\}']
\end{coreprompt}

\Needspace{6\baselineskip}
\paragraph{Causal Transition.}
\begin{coreprompt}
[Initial Observation \{t\}: '\{O\_t\}', Action \{t\}: '\{A\_t\}']

[Outcome of Action \{t\}: '\{O\_\{t+1\}\}',
Action \{t+1\}: '\{A\_\{t+1\}\}']
\end{coreprompt}

\Needspace{7\baselineskip}
\paragraph{Experience.}
\begin{coreprompt}
[Observation \{t\}: '\{O\_t\}', Action \{t\}: '\{A\_t\}']

[Experience derived from Action \{t\} -> its returned observation:
\{E\_t\}]

[Observation \{t+1\}: '\{O\_\{t+1\}\}', Action \{t+1\}: '\{A\_\{t+1\}\}']
\end{coreprompt}

\subsection{Experience Review}
\label{app:experience_review_prompt}

The full-rollout Experience condition and candidate-experience construction use the
same review interface. The input contains the environment, task, raw history, latest
action, and returned observation. Previously generated experiences are excluded.

\begin{coreprompt}
You are calibrating your own prior interaction history for the same task. Inspect the
task, the raw action-observation history, and the latest completed action and returned
observation. Return exactly one JSON object as the final answer. Use
\{"decision":"CALIBRATE","experience":"..."\} only when the latest observation
establishes a new task-relevant correction or durable progress fact that should change
how the preceding history guides later decisions in this task. Otherwise return
\{"decision":"NO\_UPDATE","experience":null\}. The experience must be one concise
sentence grounded in the latest observation. State a general implication of the
observed outcome, not a plan or a concrete next action. Do not output Markdown or
commentary around the final JSON.
\end{coreprompt}

The only valid outputs are
\texttt{\{"decision":"NO\_UPDATE","experience":null\}} and
\texttt{\{"decision":"CALIBRATE","experience":"..."\}}.

\subsection{Judging and Joint Generation}
\label{app:judge_joint_prompt_summary}

For each within-actor comparison, GPT-5.6-Sol receives the original actor prompt and two
candidate actions without actor, method, arm, or experience identifiers. Each changed
pair is judged twice with A/B order reversed; inconsistent results are ties. Identical
executable actions are ties without a judge call.

The joint Qwen3-8B actor emits exactly one calibration block followed immediately by
the environment action. Its calibration output is either
\texttt{\textless calibration\textgreater\{"decision":"NO\_UPDATE",
"experience":null\}\textless/calibration\textgreater} or a
\texttt{CALIBRATE} decision with an experience string. The initial step uses
\texttt{NO\_UPDATE}; later outputs review the immediately preceding completed
transition.

\end{document}